\documentclass[twoside]{article}

\usepackage[accepted]{aistats2021}
\usepackage[round]{natbib}

\usepackage[utf8]{inputenc} 
\usepackage[T1]{fontenc}    
\usepackage{hyperref}       
\usepackage{url}            
\usepackage{booktabs}       
\usepackage{amsfonts}       
\usepackage{nicefrac}       
\usepackage{microtype}      

\usepackage[ruled,vlined]{algorithm2e}
\usepackage{setspace}

\usepackage{epsfig,bm,subfigure,graphicx,color}
\usepackage{subfigure}
\usepackage{wrapfig}

\newcommand{\argmin}{\mathop{\mathrm{argmin\,}}}

\def\measure{\rho}

\newcommand{\boldC}{{\mathbf{C}}}
\newcommand{\boldD}{{\mathbf{D}}}

\newcommand{\boldI}{{\mathbf{I}}}

\newcommand{\boldM}{{\mathbf{M}}}

\newcommand{\boldP}{{\mathbf{P}}}
\newcommand{\boldQ}{{\mathbf{Q}}}

\newcommand{\boldW}{{\mathbf{W}}}

\newcommand{\bolda}{{\mathbf{a}}}

\newcommand{\boldp}{{\mathbf{p}}}

\usepackage{amsmath,amssymb,amsfonts,amsthm,mathtools}
\usepackage{bbm}

\newtheorem{theorem}{Theorem}

\newtheorem{proposition}[theorem]{Proposition}

\theoremstyle{definition}

\newtheorem{example}[theorem]{Example}

\newenvironment{proofpn}[1]{%
\renewcommand{\proofname}{Proof of Proposition {#1}}\proof}{\endproof}

\allowdisplaybreaks

\newcommand{\defeq}{\vcentcolon=}

\begin{document}

\twocolumn[

\aistatstitle{Fast and Robust Comparison of Probability Measures in Heterogeneous Spaces}

\aistatsauthor{ Ryoma Sato \And Marco Cuturi \And  Makoto Yamada \And Hisashi Kashima }

\aistatsaddress{ Kyoto University\\RIKEN AIP\And CREST-ENSAE\\Google Brain \And Kyoto University\\RIKEN AIP \And Kyoto University\\RIKEN AIP } ]

\begin{abstract}
Comparing two probability measures supported on heterogeneous spaces is an increasingly important problem in machine learning.
Such problems arise when comparing for instance two populations of biological cells, each described with its own set of features, or when looking at families of word embeddings trained across different corpora/languages. 
For such settings, the Gromov Wasserstein (GW) distance is often presented as the gold standard. GW is intuitive, as it quantifies whether one measure can be isomorphically mapped to the other. However, its exact computation is intractable, and most algorithms that claim to approximate it remain expensive. 
Building on~\cite{memoli-2011}, who proposed to represent each point in each distribution as the 1D distribution of its distances to all other points, we introduce in this paper the Anchor Energy (AE) and Anchor Wasserstein (AW) distances, which are respectively the energy and Wasserstein distances instantiated on such representations. Our main contribution is to propose a sweep line algorithm to compute AE \emph{exactly} in log-quadratic time, where a naive implementation would be cubic. This is quasi-linear w.r.t. the description of the problem itself.  
Our second contribution is the proposal of robust variants of AE and AW that uses rank statistics rather than the original distances. We show that AE and AW perform well in various experimental settings at a fraction of the computational cost of popular GW approximations. Code is available at \url{https://github.com/joisino/anchor-energy}.
\end{abstract}

\section{Introduction}
\label{introduction}
Wasserstein distances have proved useful to compare two probability distributions when they are both supported in the \textit{same} metric space. This is exemplified by its several applications, notably in image \citep{ni2009local,SW,de2012blue,schmitz2018wasserstein} and natural language processing \citep{kusner2015word,rolet2016fast}, neuroimaging~\citep{janati2020multi}, single-cell biology \citep{schiebinger2019optimal,yang2020predicting}, and when training generative models \citep{WGAN,salimans2018improving,SinkhornLoss}. Yet, when the two distributions live in two \textit{different} and seemingly unrelated spaces, simple Wasserstein distances fail and one must to resort to the more involved framework of Gromov-Wasserstein (GW) distances \citep{memoli-2011}. 
 
\textbf{Generality of GW.} The GW distance replaces the linear objective appearing in optimal transport (OT) with a quadratic function of the transportation plan that quantifies some form of metric distortion when transporting points from one space to another. GW is not only an elegant answer to this problem, but it is also well grounded in theory \citep{sturm2012space} and has been successfully applied to shape matching \citep{memoli07, GWSinkhorn}, machine translation \citep{melis-translation, grave-translation}, and graph matching \citep{graph-matching,SGWL}. The GW distance compares two distributions supported on spaces that are endowed with a metric, or more generally, a cost structure. This \textit{``(distribution, metric)''} pair is called a \emph{measured metric spaces} (MMS), which reduces, in a discrete setting to a probability vector of size $n$ paired with an $n \times n$ cost matrix.

\textbf{GW approximations.} Although well grounded in theory, the GW geometry is not tractable computationally: since its exact computation requires solving an NP-hard quadratic assignment problem (QAP) \citep{memoli07}, practical applications rely on approximation schemes to approach GW. 
\citet{QP} relaxed the problem into a convex quadratic program, while \citet{SDP} relaxed it to a semidefinite program. Although tractable for small $n$, these relaxations have scalability issues that prevent their use beyond a few hundred points. \citet{GWSinkhorn} and \citet{GWbary} proposed to modify the QAP using entropic regularization. When comparing two MMSs supported on $n$ points, this results in an algorithm that alternates between a local linearization of the GW objective (requiring two matrix multiplications at a $2 n^3$ cost) followed by Sinkhorn iterations (for a $O(n^2)$ cost).
This combination of outer/inner loops can prove too costly for large $n$, and can suffer from numerical instability due to the fact that the scale of the cost matrix can change at each linearization, making the Sinkhorn kernel application numerically unstable. To bring down computational costs further, \citet{SGW} proposed to restrict the GW problem to measures supported on Euclidean spaces, and to generalize the slicing approach \citep{SW} to obtain a cheaper $O(n \log n)$ computational price, pending additional efforts to rotate/register point clouds. This lower complexity if obtained by placing restrictions on the data, since it cannot handle non-Euclidean data or costs (graphs, geodesic distances of shapes, and more general arbitrary kernels and costs).

\textbf{From Points to Distributions of Costs.} An effective way to compare two points living in different MMSs could be achieved by giving them a common feature representation. Such a representation can be obtained by mapping each point from an MMS onto the 1D distribution of all its distances/costs to all other points in its MMS (hence the characterization of such points as ``anchors''). This approach was introduced in computer graphics and vision to compare 3D shapes \citep{osada2002shape, hamza2003geodesic, gelfand2005robust, manay2006integral} and linked with the GW problem by~\cite{memoli-2011}. As a result, two points from homogeneous sources can be compared through their respective cost distributions in $\mathcal{P}(\mathbb{R})$. Consequently, MMSs can be represented as elements of $\mathcal{P}(\mathcal{P}(\mathbb{R}))$ and compared with a suitable metric on that space.

\textbf{Contributions.} We focus first on the energy distance \citep{szekely2013energy,sejdinovic2013equivalence} computed between measures in $\mathcal{P}(\mathcal{P}(\mathbb{R}))$, using the 1-Wasserstein distance as a negative definite kernel to compare atoms in $\mathcal{P}(\mathbb{R})$ (Wasserstein distances are indeed n.d. in $\mathcal{P}(\mathbb{R})$, as remarked by \citet{kolouri2016sliced}). We call this approach the Anchor Energy (AE) distance, and our first contribution is to propose an efficient approach to compute it in $O(n^2 \log n)$ time, resulting in \textit{quasi-linear} complexity w.r.t. the input size (an MMS is described with an $n^2$ cost matrix). 
We also consider the entropic regularized Wasserstein distance of the distributions of anchor 1D distributions, an call this variant the Anchor Wasserstein (AW) distance. Our second contribution is to introduce a rank-based approach in which raw distance values in a MMS are replaced by their ranks, relative to \textit{all} pairwise distances described in the MMS. We show that the AE and AW distances improve on entropic GW in 3D shape comparison and node assignment tasks, while being order of magnitudes faster and applicable to larger scales.
\section{Optimal Transport and Anchors}

We call any measure $\mu$ on a measurable space $\mathcal{X}$ endowed with a cost function $\mathcal{X}\times\mathcal{X}\rightarrow \mathbb{R}$ a \emph{measured metric space} (MMS). Equivalently, a discrete measured metric set (MMSet) of size $n$ is a pair $S = (\bolda, \boldC)\in\Sigma_n\times\mathbb{R}^{n\times n}$, where, for a given $n\geq 1$, $\Sigma_n = \{\bolda \in \mathbb{R}_+^n \mid \sum_i \bolda_i = 1 \}$ is the $n$-probability simplex. Intuitively an MMSet is a probability vector $\bolda$, putting mass on points seen exclusively from the lens of an $n \times n$ pairwise cost matrix $\boldC$. Such pairs of weights/cost matrices typically arise when describing a weighted point cloud, along with the shortest path distance matrix induced from a graph on that cloud, or more generally any other cost function \citep{GWSinkhorn,GWbary}. 
We assume no knowledge on the points that constitute an MMSet, and only work from $\boldC$ (the SGW framework assumes that $\boldC$ is a squared-Euclidean distance matrix). Before describing the GW distance between two MMSets, we review the standard Wasserstein distance and quadratic assignments.

\textbf{Wasserstein Distance.}
Given two measurable spaces $\mathcal{X}$ and $\mathcal{Y}$, a cost function $c\colon\mathcal{X}\times\mathcal{Y}\rightarrow \mathbb{R}$, and two measures $\mu\in\mathcal{P}(\mathcal{X})$ and $\nu\in\mathcal{P}(\mathcal{Y})$, the optimal transport problem between $\mu$ and $\nu$ can be written as 
\begin{align*}
\text{OT}(\mu,\nu)=\min_{\gamma\in\Pi(\mu,\nu)} \mathbb{E}_{(X,Y)\sim \gamma}[c(X,Y)],
\end{align*}
where $\Pi(\mu,\nu)$ is the set of joint couplings on $\mathcal{X}\times\mathcal{Y}$ that have marginals $\mu,\nu$. When $\mathcal{X}=\mathcal{Y}$ and the cost $c$ is a metric on $\mathcal{X}$ raised to the power $p$ (with $p\geq 1$), the optimum $\text{OT}(\mu,\nu)^{1/p}$ is called the $p$-Wasserstein metric between $\mu$ and $\nu$. We consider next two subcases that are relevant for the remainder of this paper.

When $\bolda^{1} , \bolda^{2} \in \Sigma_n$ (i.e., discrete), a cost matrix $\boldC \in \mathbb{R}^{n \times n}$ suffices to instantiate the OT problem:
\begin{align*}
&\text{OT}(\bolda^{1}, \bolda^{2}) = \min_{\boldP \in U(\bolda^{1}, \bolda^{2})} \sum_{ij} \boldP_{ij} \boldC_{ij}, \text{ where } \\ & U(\bolda^{1}, \bolda^{2}) = \{ \boldP \in \mathbb{R}_+^{n \times m} \mid \boldP \mathbbm{1}_m  = \bolda^{1}, \boldP^\top \mathbbm{1}_n = \bolda^{2} \}, \end{align*} 
is the transportation polytope and $\mathbbm{1}_m$ is the $m$-vector of ones. When $\mu,\nu\in\mathcal{P}(\mathbb{R})$ and $c(x,y)=|x-y|^p$, $p\geq 1$, one has that \citep[\S2]{SantambrogioBook}
\begin{align*}
\text{OT}_p(\mu,\nu)=\int_{0}^{1}|H^{-1}_{\mu}(u)-H^{-1}_{\nu}(u)|^p\text{d}u,
\end{align*}
where $H_\mu$ is the empirical or cumulative density function (CDF) of measure $\mu$, and therefore $H^{-1}_\mu$ is its quantile function. We will write in that case $W_p=(\text{OT}_p)^{1/p}$ for the $p$-Wasserstein distance. Note further that for $p=1$, integrating differences in CDF is equal to integrating differences in quantile functions:
\begin{align} \label{eq: 1dim}
\text{OT}_1(\mu,\nu)=\int_{-\infty}^{\infty}|H_{\mu}(u)-H_{\nu}(u)|\text{d}u.
\end{align}

\textbf{Gromov Wasserstein Distance.}
The Gromov Wasserstein (GW) problem between two MMSs $(\mu,c_1), (\nu,c_2)$ is defined as follows:
\begin{align*}
\text{GW}(\mu,\nu)\!\!= \!\!\!\!\! \min_{\gamma\in\Pi(\mu,\nu)}\!\! \mathbb{E}_{\substack{(X,Y)\sim \gamma\\(X',Y')\sim \gamma}} \!\!\left[(c_1(X,Y)-c_2(X',Y'))^2\right].
\end{align*}
When instantiated on two MMSets, $S^{1} = (\bolda^{1},\boldC^{1})$ and $S^{2} = (\bolda^{2},\boldC^{2})$, this problem reduces to
\begin{align} \label{eq: GW}
\text{GW}(S^{1}, S^{2}) = \min_{\boldP \in U(\bolda^{1}, \bolda^{2})} \sum_{ijkl} \boldP_{ik} \boldP_{jl} (\boldC^{1}_{ij} - \boldC^{2}_{kl})^2.
\end{align}
As recalled in \S1, the GW distance has found success in several applications, such as shape matching, machine translation, and graph matching. However, computing it is NP-hard \citep{memoli07}, and even simply evaluating it knowing the optimal $\gamma$ beforehand would have an $O(n^3)$ computational pricetag \citep{SGW}.

\textbf{Slicing Approach.}
As mentioned above, the Wasserstein distance between two univariate distributions (namely when $\boldC_{ij} = |x_i - x_j|$ for values $x_i\in\mathbb{R}$) can be solved by computing CDFs, and therefore only requires sorting with an $O(n \log n)$ time complexity. The sliced Wasserstein distance~\citep{SW} leverages this feature by projecting measures onto random $1$-dimensional lines, to sum up these Wasserstein distances. Recently, \citet{SGW} proposed the sliced Gromov Wasserstein (SGW) distance, which exploits this idea when comparing two distributions lying in two Euclidean spaces. To be efficient, this method requires an additional ``realignment'' step which computes a linear transform of one of the input measures. The SGW approach assumes that both MMSets are embedded in $\mathbb{R}^d$, and can only consider Euclidean geometry. That method cannot, therefore, be generalized to more generic families of costs. In this paper, we do exploit the fast computation of 1D Wasserstein distances as SGW, but without relying on a projection step.

\textbf{Points as Anchors.}
To compare two MMSets, we map each point from each MMSet to the distribution of its (weighted) ground cost or distance to all other points within that MMSet. This idea was successfully used in computer graphics \citep{gelfand2005robust, memoli-2011} under the name of \textit{local distribution of distances}, spanning several successful approaches for 3D shape comparison. Specifically, the cost distribution from the anchor indexed by $i$ within $S = (\bolda, \boldC)$ is simply the 1D empirical distribution of the cost of line $i$ in $\boldC$, weighted by $\bolda$:
\[ \measure(i, \bolda, \boldC) \defeq \sum_{j = 1}^n \bolda_j \delta_{\boldC_{ij}} \in \mathcal{P}(\mathbb{R}), \]
where for $x\in\mathbb{R}$, $\delta_x$ is the Dirac mass at $x$. Notice that this can be interpreted, for continuous measures, as the push-forward $(c(x,\cdot))_{\sharp}\mu$ for an anchor $x\in\mathcal{X}$. We represent an MMSet as a distribution of local distributions of distances
\[ \mathcal{A}(\bolda, \boldC) \defeq \sum_{i = 1}^n \bolda_i \delta_{\measure(i, \bolda, \boldC)} \in \mathcal{P}(\mathcal{P}(\mathbb{R})), \]
which we call the anchor feature representation of an MMSet. Although the anchor feature map is not injective \citep{memoli-2011} (i.e., there exists two MMSets $S^1 \neq S^2$ such that $\mathcal{A}(S^1) = \mathcal{A}(S^2))$, many powerful features for 3D shape comparison, such as the global distribution of distances \citep{osada2002shape} and eccentricity \citep{hamza2003geodesic, manay2006integral}, can be computed from the anchor feature of a 3D shape. See also \citep{boutin2004reconstructing} for conditions under which a point cloud can be reconstructed from its anchor features.
\section{Anchor Distances and Plans}
\label{method}
We build on the several ideas introduced in the previous section to present several tools to handle MMSets in practice, when regarded as measures in $\mathcal{P}(\mathcal{P}(\mathbb{R}))$. We first start with a simple entropic regularization of a previous proposal by~\citep{memoli-2011}, follow with instantiating an energy distance (which we show can be computed in quadratic time in \S\ref{sec:SLAE}) and introduce a way to recover a plan from it. We conclude this section with a simple proposal to robustify our quantities that has a quantifiable experimental impact.

\textbf{Anchor Wasserstein (AW).}
The first anchor-based distances that we propose is the entropic-regularized Wasserstein distance on $\mathcal{P}(\mathcal{P}(\mathbb{R}))$ using the Wasserstein distance on $\mathcal{P}(\mathbb{R})$ as the ground metric. Notice that this ``hierarchy'' of Wasserstein distance defined using Wasserstein distances has proved useful in related tasks~\citep{yurochkin2019hierarchical, dukler2019wasserstein}. This yields in our case,
\begin{align*}
    &\text{AW}_{p}(S^1, S^2) \defeq  \text{OT}_{\varepsilon}(\mathcal{A}(S^1), \mathcal{A}(S^2)) \\ &\quad = \min_{\boldP \in U(\bolda^1, \bolda^2)} \sum_{ij} \boldP_{ij} \text{OT}_p^p(\measure_i^1, \measure_j^2) - \varepsilon H(\boldP) \\
    &\quad = \min_{\boldP \in U} \sum_{ij} \boldP_{ij} \min_{\boldQ \in U} \sum_{kl} \boldQ_{kl} ~ |\boldC^{1}_{ik} - \boldC^{2}_{jl} |^p - \varepsilon H(\boldP),
\end{align*}
where $U = U(\bolda^1, \bolda^2)$ is the set of transportation plans, $\measure_i^k = \measure(i, \bolda^k, \boldC^k)$ is the local distribution of distances of the $i$-th node in either the first or second MMSet, $H(\boldP) = - \sum_{ij} \boldP_{ij} (\log \boldP_{ij} - 1)$ is the entropy, and $\varepsilon > 0$ is the regularization coefficient. This formula is similar to that of the GW distance (Eq. \ref{eq: GW}), but AW optimizes global and local assignments separately.
In the case of $\varepsilon = 0$, this is equal to the lower bound (TLB) of the GW distance introduced in \citep{memoli-2011}.
However, $\varepsilon = 0$ requires more computational cost for solving a linear programming \citep{memoli-2011}. Thanks to the entropic regularization, the Sinkhorn algorithm \citep{lightspeed} can speed up the computation. Namely, the cost matrix $\boldC_{ij} = OT_p^p(\measure_i^1, \measure_j^2)$ can be computed in $O(nm(n+m))$ time, and the OT matrix $\boldP^*$ can be computed in quadratic time by the Sinkhorn algorithm. The total time complexity is cubic, but unlike GW the cost matrix is computed only once and the energy is convex. This cubic computational complexity might be, however, prohibitive in some applications. We switch to a simpler tool to reach a quasi-quadratic complexity.

\textbf{Anchor Energy (AE).}
The second variant of the anchor-based distances is energy distance~\citep{szekely2013energy} using the 1D Wasserstein kernel. Specifically,
\begin{align} \label{eq: AE_def}
\text{AE}_p(S^1, S^2) &\defeq 2 \mathbb{E}_{\substack{h \sim \mathcal{A}(S^1) \\ g \sim \mathcal{A}(S^2)}}[\text{OT}_p^p(h, g)] \notag \\ &\quad - \mathbb{E}_{\substack{h_1 \sim \mathcal{A}(S^1) \\ h_2 \sim \mathcal{A}(S^1)}}[\text{OT}_p^p(h_1, h_2)] \\ &\quad - \mathbb{E}_{\substack{g_1 \sim \mathcal{A}(S^2) \\ g_2 \sim \mathcal{A}(S^2)}}[\text{OT}_p^p(g_1, g_2)]. \notag
\end{align}
$\text{OT}_1$ and $\text{OT}_2^2$ are conditionally negative definite because turning a measure into a quantile function is a feature map onto a Hilbert space of functions \citep{kolouri2016sliced}. Therefore, $\text{AE}_p$ is a valid energy distance for $p = 1, 2$ \citep{sejdinovic2013equivalence} and provides a metric structure to the anchor feature space. In particular, this is a pseudometric in the original measure space. Note that the energy distance can be interpreted as maximum mean discrepancy (MMD) via the distance-induced kernel \citep{sejdinovic2013equivalence}. One can also see that Eq. \ref{eq: AE_def} corresponds to the MMD for comparing graph generative models \citep{you2018graphrnn} when the anchor feature is the distribution of node feature distributions. Thus our proposed algorithm can compare graph generative models efficiently as well as measures in heterogeneous spaces. We confirm this by statistical tests of graph generative models in the experiments.

Each term of Eq. \ref{eq: AE_def} can be explicitly expressed as follows:
\begin{align} \label{eq: AE}
    &\mathbb{E}_{\substack{h \sim \mathcal{A}(S^1) \\ g \sim \mathcal{A}(S^2)}}[\text{OT}_p^p(h, g)] \notag = \sum_{ij} \bolda^1_i \bolda^2_j ~\text{OT}_p^p(h, g) \notag \\ &= \sum_{ij} \bolda^1_i \bolda^2_j \min_{\boldP \in U(\bolda^1 \bolda^2)} \sum_{kl} \boldP_{kl} |\boldC^1_{ik} - \boldC^2_{jl}|^p.
\end{align}
The last representation looks similar to the GW distance (Eq. \ref{eq: GW}), but the AE distance calculates an average of the local OT of all pairs of points, while the GW distance computes a global OT. The AE distance can be seen as the simplified version of the AW distance (thus of the TLB \citep{memoli-2011}) that fixes the local global assignment to the marginalized transportation. \emph{The AE distance requires cubic time when computed naively} because computing an OT of sorted 1D distributions takes linear time. We show below that the AE distance can be, in fact, computed in {quasi-quadratic} $O((n^2 + m^2) \log (nm))$ time in Section~\ref{sec:SLAE}. We highlight two other contributions in this section: the AE distance can also be used to devise an efficient transportation plan, and anchor distances can be ``robustified'' by replacing costs in $\mathbf{C}$ by their ranks.

\textbf{Anchor Energy Plan (AEP).}
A strong appeal of OT and GW-based distances is that they also provide assignments. We propose the AE plan (AEP) to fill in that gap,
\begin{align*}
    \textbf{AEP}(S^1, S^2) = \sum_{ij} \bolda^1_i \bolda^2_j \argmin_{\boldP \in U(\bolda^1 \bolda^2)} \sum_{kl} \boldP_{kl} |\boldC^1_{ik} - \boldC^2_{jl}|^p.
\end{align*}
This assignment takes an average of the local OTs of all pairs of points as the AE distance, while the assignment by the GW distance computes a global OT. Therefore, AEP cannot distinguish detailed structure or take interactions between points into consideration, but AEP can provide robust assignment that preserves the role of points. For example, suppose there exists an outlier point in an MMSet. The GW assignment is influenced by this outlier point. In contrast, a small fraction of local OTs are affected by this outlier, and affected transportation plans do not take too large values because all transportation plans must be doubly-stochastic. Therefore, transportation plans of inlier points are dominant terms in AEP. We empirically confirm that AEP can provide a good assignment that preserves roles of nodes in networks in the experiments. AEP can be computed in $O(nm(n+m) + n^2 \log n + m^2 \log m)$ time, and an assignment of a single node is computed in $O(nm \log(nm) + m^2 \log m)$ time by naive computation. 

\textbf{Robust Anchor Representation.} All of our algorithms operate by considering input values $\mathbf{C}^1$ and $\mathbf{C}^2$. Because these values may take very different ranges, the problem of rescaling in a suitable way is crucial in most applications. Yet, the anchor features we have introduced are not robust to scaling since $\mathcal{A}(\bolda, \boldC)$ will likely differ from $\mathcal{A}(\bolda, \lambda \boldC)$ when $\lambda \neq 1$.  This issue is a common one when using GW distances. Although dividing the distance by the largest value in the distance matrix can alleviate this problem, if anomaly points, noise, or clutters are contained in the MMSets, for example due to the scanning process of 3D shapes, it is difficult to settle this problem. To overcome this issue, we propose a robust variant of the anchor feature. Instead of using matrix $\boldC$ directly, we pre-process them to output their normalized rank-based statistics:

$$\widetilde{\mathbf{C}} = \frac{1}{n^2} \text{reshape}\left(\sigma_{\text{argsort}}(\text{vec}(\mathbf{C})\right), (n,n)).$$

This translates into rank-based anchor feature distributions (Appendix \ref{sec: robust}). The rank-based anchor features can be computed through a simple argsort of the cost matrices, in $O(n^2 \log n)$ time. Our experiments show that the distances using the robust anchor feature perform better when the scales of the entries in cost matrices of MMSets differ significantly. We also confirm in the appendix that this modification also ``robustifies'' GW.

\section{Sweep Line Anchor Energy.}\label{sec:SLAE}
\begin{figure}
    \centering
    \includegraphics[width=\hsize]{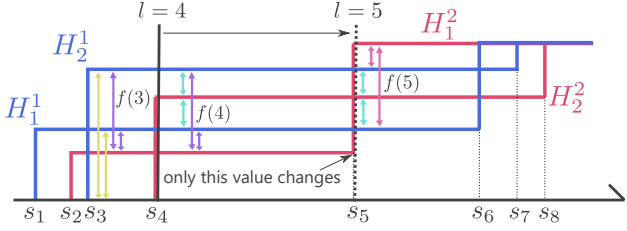}
    \caption{\textbf{Overview of SL-AE:} The AE distance is the integration of the total variations $f(l)$ of CDFs. SL-AE sums up the total variations $f(1), f(2), \dots, f(n^2 + m^2-1)$ sweeping a vertical line from left to right. The key idea is only terms that involve a single CDF change in each segment, and these total variations can be managed efficiently by balanced trees.}
    \label{fig: SL-AE}
\end{figure}
In this section, we propose an efficient computation algorithm for the AE distance, using the sweep line algorithm. We refer to this quantity as Sweep Line Anchor Energy (SL-AE), but one should keep in mind that AE and SL-AE are equal, and that AE refers to a naive summation approach that has cubic time, while SL-AE refers to a mover evolved approach to compute the same quantity. SL-AE computes the AE distance in $O((n^2 + m^2) \log (nm))$ time for any integer $p \geq 1$. We introduce the case of $p = 1$ for simplicity, but the idea can be naturally extended to general integers $p$. SL-AE uses a technique inspired from the sweep line algorithm \citep{shamos1976geometric, fortune1987sweepline}, which was originally proposed in the computational geometry field. Figure \ref{fig: SL-AE} illustrates the algorithm behind SL-AE.

\textbf{Summary of SL-AE:} The AE distance requires at least cubic time if each OT is computed separately, because the AE distance involves $nm$ OT evaluations, and each OT evaluation takes at least linear time. SL-AE computes the summation of all OT distances at once to speed up computations. There are $n$ ``red'' CDFs from the first distribution and $m$ ``blue'' CDFs from the second distribution. The AE distance is the sum of areas between all $nm$ pairs of red and blue functions because the $1$-Wasserstein distance of 1D distributions is equal to the area between the CDFs of the distributions (Eq. \ref{eq: 1dim}). SL-AE integrates these areas from left to right at once rather than computing each area separately. This change of viewpoint is important, yet another important insight is that each CDF is piecewise constant, and the CDF of $\measure(i, \bolda^1, \boldC^1)$ changes at only $n$ points $\boldC_{i1}, \boldC_{i2}, \dots, \boldC_{in}$, which makes $L = n^2 + m^2$ change points $s_1, \dots, s_L$ in total. We assume all change points are unique without loss of generality. SL-AE cuts the real line at these change points, which makes $(n^2 + m^2 - 1)$ segments. In each segment, all CDFs are constant. Thus, the sum of the areas in each segment $l$ is the total variation $f(l)$ of red and blue CDFs multiplied by the length $(s_{l+1} - s_l)$ of the segment. In addition, adjacent segments are almost the same, and only one CDF changes. Although the difference $(f(l+1) - f(l))$ involves $O(n + m)$ terms, efficient data structures, such as B-tree and Fenwick tree, offer a logarithmic time algorithm to compute the partial summation. In total, there are $(n^2 + m^2 - 1)$ segments, and it takes logarithmic time to update the total variation in each segment; thus, the time complexity of SL-AE is $O((n^2 + m^2) \log (nm))$.

We can now describe formally our algorithm. Let $L = n^2 + m^2$ be the number of elements in the cost matrices and $(i_l, j_l, k_l) \in \{1, \dots, n\} \times \{1, \dots, m\} \times \{1, 2\} ~(l = 1, 2, \dots, L)$ be indices of the cost matrices in increasing order of their values. Thus $\boldC^{k_1}_{i_1 j_1} \le \boldC^{k_2}_{i_2 j_2} \le \dots \le \boldC^{k_L}_{i_L j_L}$. Let $s_l = \boldC^{k_l}_{i_l j_l} ~(l = 1, \dots, L)$ and let $H^{k}_i(x)$ be the CDF of $\measure(i, \bolda^k, \boldC^k)$.
\begin{proposition} \label{thm: variation}
\begin{align*}
&\mathbb{E}_{h^1 \sim \mathcal{A}(S^1), h^2 \sim \mathcal{A}(S^2)}[\text{OT}_1(h^1, h^2)] \\ &\qquad = \sum_{l = 1}^{n^2 + m^2-1} (s_{l+1} - s_l) f(l),
\end{align*}
where $f(l) = \sum_{i = 1}^n \sum_{j = 1}^m \bolda^{1}_i \bolda^{2}_j ~|H^{1}_i(s_l) - H^{2}_j(s_l)|$ is the total variation in $x \in [s_l, s_{l+1}]$.
\end{proposition}
The key idea of SL-AE is hat in each iteration $l$, only $H^{k_l}_{i_l}$ changes (i.e., $H^{k}_i(s_{l+1}) = H^{k}_i(s_l)$ for $i \neq i_l$ or $k \neq k_l$). Let $\mathcal{S}^{k}(u, v)$ and $\mathcal{T}^{k}(u, v)$ be the partial sums of weights and weighted CDFs:
\begin{align} \label{eq: ST}
&\mathcal{S}_l^{k}(u, v) = \sum_{i\colon u \le H^{k}(i, s_l) < v} \bolda^{k}_i, \\ \label{eq: ST2}  &\mathcal{T}_l^{k}(u, v) = \sum_{i\colon u \le H^{k}(i, s_l) < v} \bolda^{k}_i H^{k}_i(s_l).
\end{align}
\begin{proposition} \label{thm: update}
\begin{align*}
f(l) = f(l-1) &- \bolda^{k_l}_{i_l} (\mathcal{S}_l^{k_l'}(-\infty, c) \cdot c - \mathcal{T}_l^{k_l'}(-\infty, c)) \\ &- \bolda^{k_l}_{i_l} (\mathcal{T}_l^{k_l'}(c, \infty) - \mathcal{S}_l^{k_l'}(c, \infty) \cdot c) \\
&+ \bolda^{k_l}_{i_l} (\mathcal{S}_{l+1}^{k_l'}(-\infty, c') \cdot c' - \mathcal{T}_{l+1}^{k_l'}(-\infty, c')) \\ &+ \bolda^{k_l}_{i_l}(\mathcal{T}_{l+1}^{k_l'}(c', \infty) - \mathcal{S}_{l+1}^{k_l'}(c', \infty) \cdot c'),
\end{align*}
where $c = H^{k_l}_{i_l}(s_l)$ and $c' = H^{k_l}_{i_l}(s_{l+1})$, and $k'_l$ is the opposite index of $k_l$ (i.e., $k'_l = 2$ if $k_l = 1$ and $k'_l = 1$ if $k_l = 2$). $\infty$ denotes a sufficiently large constant.
\end{proposition}
We use balanced trees, such as the B-tree \citep{Btree} and Fenwick tree \citep{Fenwick}, to compute $\mathcal{S}^{k}$ and $\mathcal{T}^{k}$. A balanced tree $T$ maintains an array and computes the following operations in $O(\log n)$ time, where $n$ is the number of elements in the array.

\setlength{\leftmargini}{.3in}
\begin{itemize}
    \item $\text{Add}(T, i, x)$: Add $x$ to $T_i$.
    \item $T(u, v)$: Calculate $\sum_{i\colon u \le i \le v} T_i$.
\end{itemize}

\begin{algorithm}[tb]
\label{algo: SL-AE}
\caption{Sweep Line Anchor Energy}
\SetAlgoLined
\DontPrintSemicolon
Initialize $W$, $f$, $\mathcal{S}_1$, and $\mathcal{T}_1$ to zero.\\
Sort the segments in increasing order.\\
\For{$l = 1, 2, \dots, L-1$}{ 
Subtract the previous values (i.e., the first line in Prop. \ref{thm: update}) from $f$ by querying $\mathcal{S}_l$ and $\mathcal{T}_l$.\\
Update $\mathcal{S}_l$ and $\mathcal{T}_l$ to obtain $\mathcal{S}_{l+1}$ and $\mathcal{T}_{l+1}$ based on Eq. \ref{eq: ST} and \ref{eq: ST2}.\\
Add the next values (i.e., the second line in Prop. \ref{thm: update}) to $f$ by querying $\mathcal{S}_{l+1}$ and $\mathcal{T}_{l+1}$.\\
$W \leftarrow W + (s_{l+1} - s_{l}) \cdot f$
}
\textbf{return} $W$
\end{algorithm}

Note that an index $i$ is not necessarily an integer but a real value in general. The update of $f$ in Proposition $\ref{thm: update}$ requires $O(n + m)$ time naively because $\mathcal{S}^{k}$ and $\mathcal{T}^{k}$ involve $O(n + m)$ terms. This summation can be sped up to $O(\log (nm))$ time by balanced trees. The pseudo code of SL-AE is shown in Algorithm \ref{algo: SL-AE}, and more details are described in the Appendix.

\textbf{Time Complexity:} We analyze the complexity of SL-AE briefly. The sorting of $\{s_i\}_{i = 1, \dots n^2 + m^2}$ requires $O((n^2 + m^2) \log (nm))$ time. The loop iterates $n^2 + m^2 - 1$ times. In each iteration, there are constant numbers of updates and queries for $\mathcal{T}_i$ and $\mathcal{S}_i$. If we use balanced trees as the data structures for $\mathcal{T}_i$ and $\mathcal{S}_i$, each query and update requires $O(\log (nm))$ times. Therefore, the total time complexity is $O((n^2 + m^2) \log (nm))$.

\section{Experiments}
\label{experiments}

We experimentally assess the performances of the AE and AW distances by answering the following questions. (Q1) \textbf{Scalability}: How fast is SL-AE compared to a naive implementation of AE? (Q2) \textbf{Trade-off}: Do AE and AW provide good trade-offs between speed and performance? (Q3) \textbf{Robustness}: Are robust versions of the AE and AW distances robust against scaling and noise? (Q4) \textbf{Assignment}: Does AEP preserve the role of nodes? (Q5) \textbf{Graph Model Comparison}: Is SL-AE useful for testing graph distributions? We also conduct experiments to compute barycenters w.r.t. the AE and AW distances in Appendix \ref{sec: barycenter}. We use the Sinkhorn algorithm for the GW distance with entropic regularization \citep{GWSinkhorn} as a baseline method (denoted by GW). We run the Sinkhorn algorithm in the log space for numerical stability. The experimental details are provided in Appendix \ref{sec: experimental_details}.

\begin{figure}[tb]
\centering
\includegraphics[width=0.7\hsize]{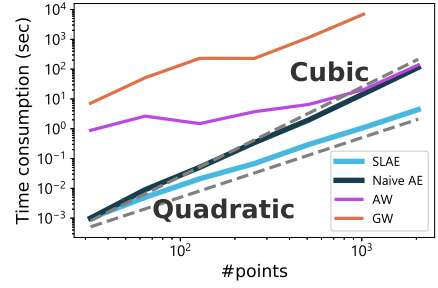}
\vspace{-0.1in}
\caption{Scalability w.r.t. number of points of various distances for MMSets, using cat shapes from Fig.~\ref{fig: shapes}.}
\label{fig: scalability}
\end{figure}
\begin{figure}[t]
    \begin{center}
    \includegraphics[width=0.8\hsize]{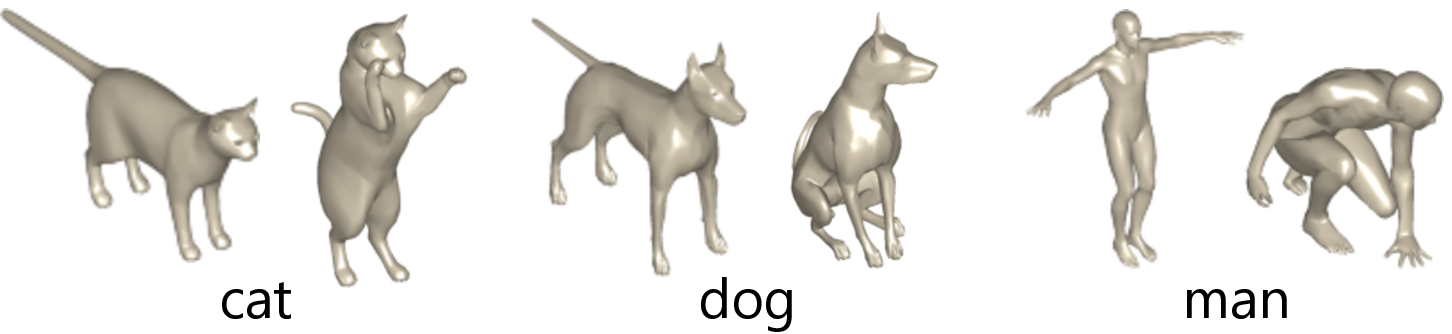}
    \includegraphics[width=0.8\hsize]{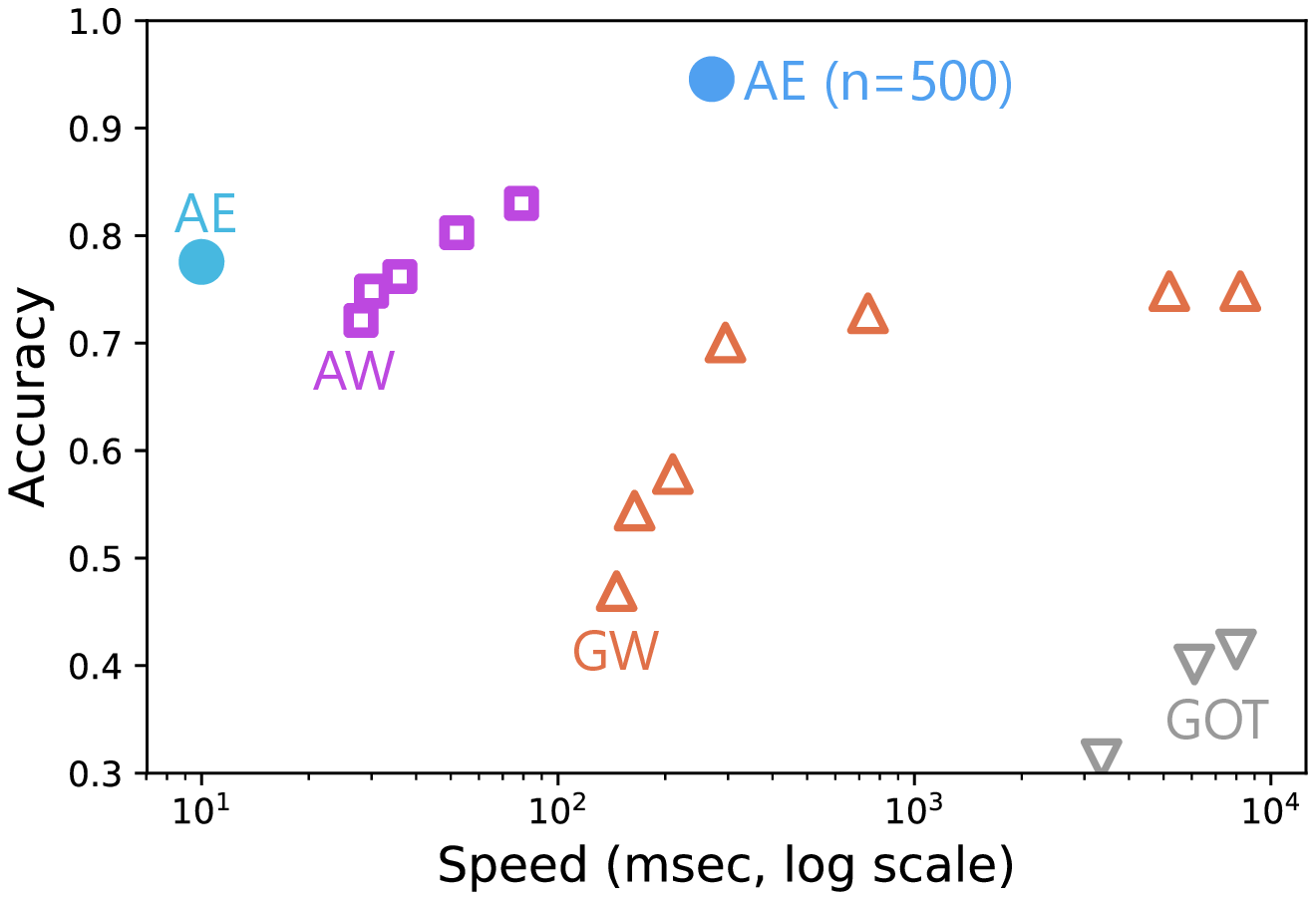}
    \end{center}
    \vspace{-0.2in}
    \caption{\textbf{3D Shape Comparison:} (Top) Examples of 3D shapes. (Bottom) Trade-off between the speed and performance of each method.}
    \label{fig: shapes}
\end{figure}

\textbf{Scalability.} We assess the scalability of our linear sweep approach. We sample $n = 32, 64, ..., 2048$ points from the 3D cat shapes illustrated in Figure \ref{fig: shapes} (Top) and compute the AE distance using the naive triple loop method and SL-AE with only a \emph{single} core of a CPU, which is desirable to compute many pairs of shapes in parallel. We also compute the AW distance and GW distance for illustrative purposes. Figure \ref{fig: scalability} gives the orders of magnitude to execute each method, highlighting the quasi-quadratic complexity of SL-AE.

\begin{figure}[t]
    \begin{center}
    \includegraphics[width=0.95\hsize]{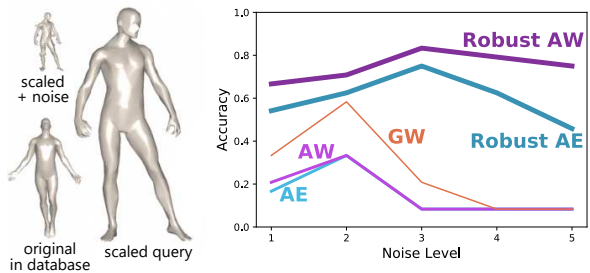}
    \end{center}
    \vspace{-0.1in}
    \caption{\textbf{Robust 3D shape comparison:} (Left) Examples of original shapes as well as scaled and scaled + noisy queries. (Right) Accuracy of $1$-NN classification using each distance.}
    \label{fig: SHREC10}
\end{figure}

\begin{figure}[t]

    \centering
    \includegraphics[width=0.9\hsize]{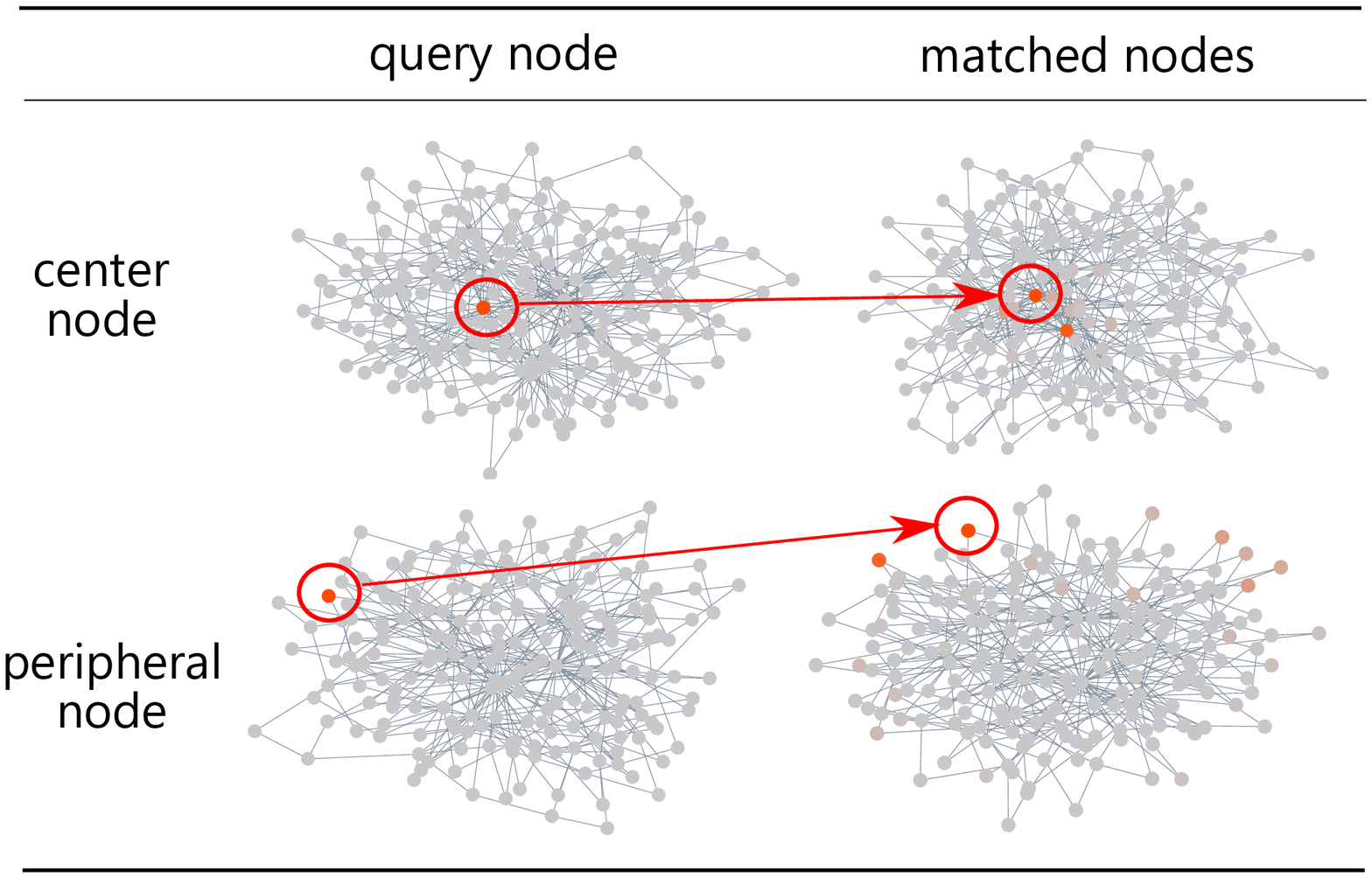}
    \includegraphics[width=0.9\hsize]{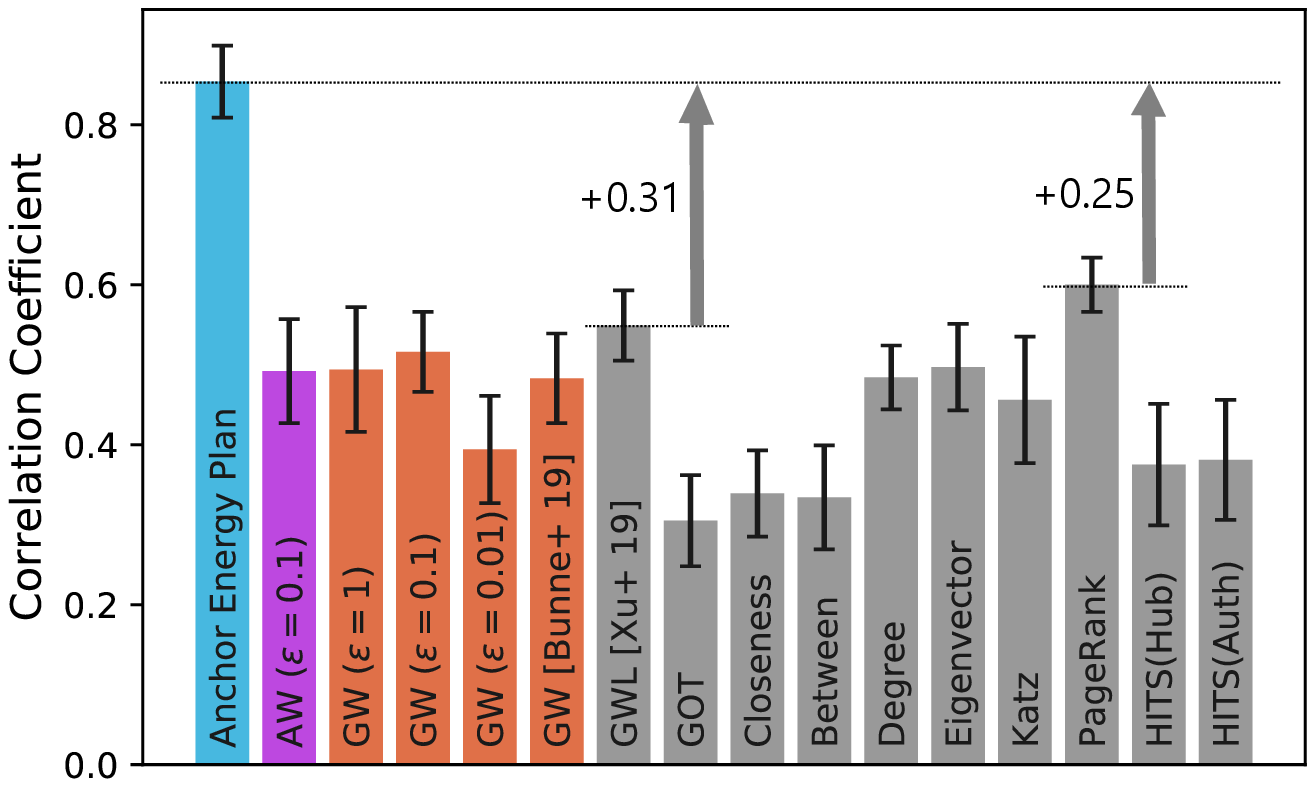}
    \vspace{-0.2in}
    \caption{\textbf{Node Assignment:} (Top) Red nodes in the right indicate that the nodes are matched, and the gray nodes in the right indicate that the nodes are not matched. This shows that the center node matches with the center nodes and the peripheral node matches with the peripheral nodes. (Bottom) Correlation coefficients of the node indices of the matched nodes. }
    \label{fig: matching}
 \vspace{-0.2in}
\end{figure}

\textbf{Trade-off.}
We assess the trade-off between the speed and performance of the AE and AW distances via 3D shape comparisons. As the regularization coefficient $\varepsilon$ tends to zero, the number of Sinkhorn iterations needed for AW and GW to converge grows. Balancing a good trade-off between the speed and performance is important in practice. We use the non-rigid world dataset \citep{3Ddata}, which contains $148$ 3D shapes of $12$ classes, to assess the trade-off between the performance and speed. Figure \ref{fig: shapes} (Top) shows examples of 3D shapes. We use the uniform weights and geodesic distances on 3D shapes as the cost function. We randomly sample $n = 100$ vertices to construct cost matrices so that the GW distance is feasible, although the SL-AE algorithm works efficiently even for larger shapes. $148 \cdot 147 / 2 = 10878$ pairs of shapes are compared in total to conduct $1$-NN classification. Figure \ref{fig: shapes} (Bottom) shows the Pareto optimal points of the $1$-NN classification accuracy and average speed to compare a single pair of shapes for each distance when we vary the regularization coefficient $\varepsilon$. Each point represents one configuration of hyperparameters. This figure also reports the accuracy of the AE distance with $n = 500$ points and GOT \citep{maretic2019GOT} for illustrative purposes. This result highlights that the AE distance performs comparably to the AW and GW distances even though the AE distance works fast and has no hyperparameters.

\textbf{Robustness.}
We apply the robust versions of the AE and AW distances to the SHREC10 robustness benchmark \citep{SHREC10}, which contains shapes in the database and transformed query shapes for each shape. We use scaling transformation queries and scaling + noise transformation queries in this experiment. Each noised query has a noise level from $1$ (weak) to $5$ (strong) in this dataset.  We use the normal and robust versions of the AE and AW distances to retrieve a shape in the database corresponding to the query shape by the nearest neighbor retrieval. We use the training data, which contain 12 shapes and 120 query shapes, for evaluation because no training phase is involved in our classifier. Figure \ref{fig: SHREC10} (Right) reports the accuracy for each noise level as well as the accuracy of the GW distance as a reference record. This indicates that the robust versions of the AE and AW distances can handle the difference of scaling, whereas normal distances suffer from a drop in accuracy as the noise level increases. We confirm that our rank-based statistics can make GW robust as well in Appendix \ref{sec: barycenter}.

\textbf{Assignment.} 
In this experiment, we assess performance on the node assignment task. We use the Barabasi-Albert (BA) model \citep{BAmodel} in this task. The BA graphs resemble real-world networks, such as citations networks and social networks, thanks to its preferential attachment mechanism. The latent order of each node in the BA model represents the order in which the node participates in the network. Therefore, smaller order nodes play more central roles than larger order nodes. We compute node assignments of two BA graphs using a variety of graph matching algorithms. A good matching algorithm is expected to match each node with a node with a similar latent order (i.e., similar role). We compute the correlation coefficient between the orders of matched nodes. High correlation coefficients indicate that the assignment preserves the role of nodes. 

We compute the AEP, GW matching, and optimal global assignment $\boldP^*$ of the AW distance. We use an open implementation \citep{GWGAN} to tune hyperparameters for the GW matching in addition to various hyperparameters. We also use GWL \citep{graph-matching} and GOT \citep{maretic2019GOT}, which are state-of-the-art OT-based methods for graph matching, and the various centrality-based assignments, where nodes are matched in order of centrality measure. We use the geodesic distances as the ground metric in this task. The examples of AEP in Figure \ref{fig: matching} (Top) show that the center node matches with center nodes and the peripheral node matches with peripheral nodes. Figure \ref{fig: matching} (Bottom) reports the average correlation coefficients of ten pairs of graphs. This indicates that the AEP preserves the roles of nodes better than the other methods. The GW and AW matching fail this task because they keenly fit noise signal, while the AEP is robust to noise thanks to averaging transportation plans.

\begin{figure}[tb]
    \centering
    \includegraphics[width=0.7\hsize]{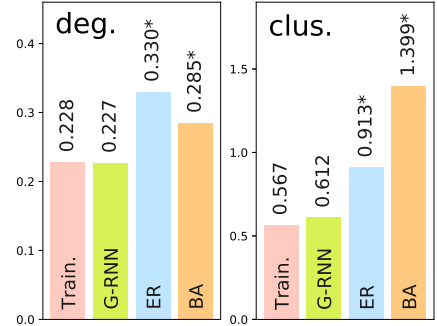}
    \vspace{-0.2in}
    \caption{Energy distance between the test dataset and each model. $^*$ means the distance is significant.}
    \label{fig: test}
    \vspace{-0.2in}
\end{figure}

\textbf{Graph Generative Model Comparison.}
Our proposed method can \emph{evaluate} graph generative models efficiently. A graph generative model receives a seed and outputs a graph. The statistics of the generated graphs, such as the degree and clustering coefficient distributions, of a good generative model should resemble that of the real graphs. We compare graph distributions using energy distance with the Wasserstein kernel of the degree and clustering coefficient distributions, following \citep{you2018graphrnn}. However, naive computation of the energy distance is time consuming due to its $O(s^2 n)$ complexity, where $s$ is the number of samples and $n$ is the number of nodes. Our proposed method enables \emph{exact} computation of the energy distance in $O(sn \log (sn))$ time. Thus our method is efficient when the number of samples is large. We use the ENZYMES dataset \citep{schomburg2004brenda, borgwardt2005protein} to train a GraphRNN \citep{you2018graphrnn}, Erd\H{o}s-R\'{e}nyi (ER) model \citep{ERmodel}, and BA model. We then compute the energy distance between the test dataset and training dataset, GraphRNN, ER, and BA model using our proposed method and conduct two-sample tests by the permutation test with $\alpha = 0.05$. Figure \ref{fig: test} shows that the weak baselines (i.e., ER and BA) are rejected while the GraphRNN model is not rejected. In other words, the GraphRNN model resembles the real distribution in terms of the energy distance, while the ER and BA models are far from the real distribution. This result indicates that our propose method can detect the effectiveness of GraphRNN against baselines. This result is consistent with \cite{you2018graphrnn}.

\section{Conclusion}
\label{conclusion}
We introduced in this paper several tools to compared two measured metric sets lying in heterogeneous spaces. Our approaches build on the anchor features of both measures to represent these measures, which are originally lying in different spaces, in the same space of probability measures on probability measures on the line. The AE distance is the energy distance and AW distance is the entropic-regularized Wasserstein distance of the anchor features. We proposed an exact algorithm to compute the AE distance exactly in $O((n^2 + m^2) \log (nm))$ time, shaving a linear term w.r.t a naive implementation. We also proposed AEP, a novel assignment method inspired by the AE distance. We experimentally showed that our sweeping line approach scales quasi-quadratically w.r.t. the size of the support, and it is faster than other OT-based metrics. We also showed that our proposed methods perform well in shape comparison and node assignment tasks, leveraging a simple robustified variant of these methods. While AE and AW distances cannot claim to capture structural correspondences at the level that can be reached by the GW framework, they do offer stable (convex) and efficient numerical approaches to reach the same goal. We argue that for some tasks, such shape comparison or social network analysis, this tradeoff is favorable to AE distance, which reaches comparable performance to that of GW with simpler computations and concepts. Code is available at \url{https://github.com/joisino/anchor-energy}.

\bibliography{cite}
\bibliographystyle{plainnat}
\appendix
\clearpage
\onecolumn
\setlength{\textfloatsep}{0.1in}
\begin{algorithm}[tb]
\label{algo: SL-AE_detailed}
\caption{Detailed pseudo code of SL-AE.}
\SetAlgoLined
\DontPrintSemicolon
 $W \leftarrow 0$, $f[0] \leftarrow 0$\\
 $H[0][i] \leftarrow 0 ~(i = 0, \dots, n)$\\
 $H[1][j] \leftarrow 0 ~(j = 0, \dots, m)$\\
 $t[i \cdot n + j] \leftarrow (\boldC^{1}_{ij}, i, j, 1, 2) ~ (i, j = 1, \dots, n)$\\
 $t[n^2 + i \cdot n + j] \leftarrow (\boldC^{2}_{ij}, i, j, 2, 1) ~ (i, j = 1, \dots, m)$\\
 $(s_l, i_l, j_l, k_l, k'_l) \leftarrow \text{sorted}(t)[l] ~(l = 1, \dots, n^2 + m^2)$\\
 Initialize $\mathcal{S}^{1}$, $\mathcal{S}^{2}$, $\mathcal{T}^{1}$, and $\mathcal{T}^{2}$ to $0$\\
 \For{$l = 1, \dots n^2 + m^2 -1$}{ 
    $c \leftarrow H[k_l][i_l]$\\
    $f[l] \leftarrow f[l-1]$\\
    $f[l] \leftarrow f[l] - \bolda^{k_l}_{i_l} \cdot (\mathcal{S}^{k_l'}(-\infty, c) \cdot c - \mathcal{T}^{k_l'}(-\infty, c))$\\
    $f[l] \leftarrow f[l] - \bolda^{k_l}_{i_l} (\mathcal{T}^{k_l'}(c, \infty) - \mathcal{S}^{k_l'}(c, \infty) \cdot c)$\\
    Add($\mathcal{S}^{k_l}, H[k_l][i_l], - \bolda^{k_l}_{i_l}$)\\
    Add($\mathcal{T}^{k_l}, H[k_l][i_l], - \bolda^{k_l}_{i_l} H[k_l][i_l]$)\\
    $H[k_l][i_l] \leftarrow H[k_l][i_l] + \bolda^{k_l}_{j_l}$\\
    Add($\mathcal{S}^{k_l}, H[k_l][i_l],  \bolda^{k_l}_{i_l}$)\\
    Add($\mathcal{T}^{k_l}, H[k_l][i_l],  \bolda^{k_l}_{i_l} H[k_l][i_l]$)\\
    $c' \leftarrow H[k_l][i_l]$\\
    $f[l] \leftarrow f[l] + \bolda^{k_l}_{i_l} \cdot (\mathcal{S}^{k_l'}(-\infty, c') \cdot c' - \mathcal{T}^{k_l'}(-\infty, c'))$\\
    $f[l] \leftarrow f[l] + \bolda^{k_l}_{i_l} (\mathcal{T}^{k_l'}(c', \infty) - \mathcal{S}^{k_l'}(c', \infty) \cdot c')$\\
    $W \leftarrow W + (s_{l+1} - s_{l}) \cdot f[l]$
  }
  \textbf{return} $W$
\end{algorithm}

\section{Robust Anchor Feature} \label{sec: robust}

The robust anchor feature is defined as:
\begin{align*}
p(i, j, \boldC) &\defeq \frac{1}{n^2} \# \{ (k, l) \mid \boldC_{kl} < \boldC_{ij} \} \in [0, 1], \\
\measure_R(i, \bolda, \boldC) &\defeq \sum_{j = 1}^n \bolda_j \delta_{p(i, j, \boldC)} \in \mathcal{P}([0, 1]),
\\
\mathcal{A}_R(\bolda, \boldC) &\defeq \sum_{i = 1}^n \bolda_i \delta_{\measure_R(i, \bolda, \boldC)} \in \mathcal{P}(\mathcal{P}([0, 1])),
\end{align*}
where $\#$ denotes the number of elements in the set.

\begin{example}
When the cost matrix is
\begin{align*}
    \boldC = \begin{pmatrix}
    0 & 4.5 & 4.1 & 4.3 \\
    4.5 & 0 & 0.5 & 0.4 \\
    4.1 & 0.5 & 0 & 0.5 \\
    4.3 & 0.4 & 0.5 & 0 \\
    \end{pmatrix},
\end{align*}
then the robust anchor feature corresponds to the standard anchor feature with the cost matrix
\begin{align*}
    \boldC' = \begin{pmatrix}
    0 & 15/16 & 11/16 & 13/16 \\
    15/16 & 0 & 7/16 & 5/16 \\
    11/16 & 7/16 & 0 & 7/16 \\
    13/16 & 5/16 & 7/16 & 0 \\
    \end{pmatrix}.
\end{align*}
\end{example}

\section{Experimental Details} \label{sec: experimental_details}

We use the Fenwick trees \citep{Fenwick} for the implementation of $\mathcal{S}^{k}$ and $\mathcal{T}^{k}$. We use the Sinkhorn algorithm with entropic regularization proposed by \citet{GWSinkhorn} (Algorithm 1 in \citep{GWSinkhorn}) to compute the GW distance. Note that \citet{GWSinkhorn} used $\alpha$ for the symbol of the regularization coefficient, but we use $\varepsilon$ instead. We set the momentum term to $\eta = 1$ because we found this did not affect the result much in the initial experiments. The Sinkhorn algorithm for the AW distance and the inner loop of the GW distance stops when the relative change in the transportation plan becomes less than $10^{-6}$. The outer loop of the GW distance stops when the relative change in the transportation plan or the objective value is less than $10^{-6}$. We measure the speeds of algorithms with a single core of an Intel Xeon E7-4830 CPU.

\subsection{Scalability (Q1)}

We compute the distances with the geodesic distance costs and uniform distribution. Specifically, we first compute the geodesic distance matrices $\boldD^1$ and $\boldD^2$ of the two cat shapes. For $n = 32, 64,  \dots, 2048$, we sample $n$ points uniformly and randomly for each shape and sample the corresponding rows and columns of the distance matrix to make $\boldC^1_n$ and $\boldC^2_n \in \mathbb{R}_+^{n \times n}$. We compute the AE, AW, and GW distances of $\boldC^1_n$ and $\boldC^2_n$ with the uniform distribution. We do not include the time consumption of preprocessing (i.e., the distance calculation and sampling) in the result. We set $\varepsilon = 10^{-5}$ for the AW distance and $\varepsilon = 10$ for the GW distance. The regularization constants are different because they need to adjust to the scale of the ground costs used within AW and GW, which are themselves different (one uses W1, the other the original ground metric).

\begin{figure}[t]
 \begin{minipage}{0.5\hsize}
  \begin{center}
    \includegraphics[width=\hsize]{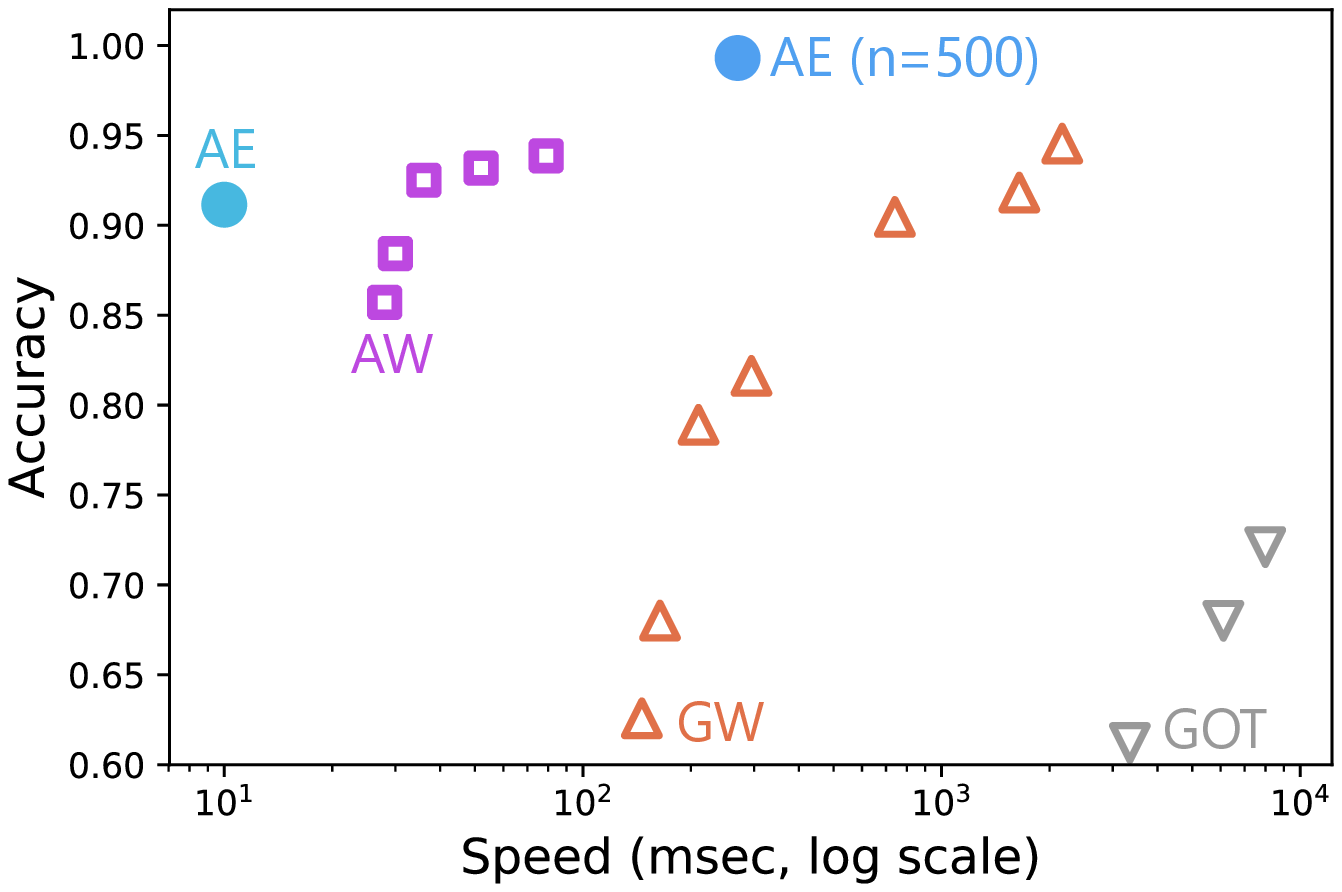}
  \end{center}
 \caption{Top-$3$ accuracy of 1-NN.}
 \label{fig: top3}
 \end{minipage}
 \begin{minipage}{0.5\hsize}
  \begin{center}
    \includegraphics[width=\hsize]{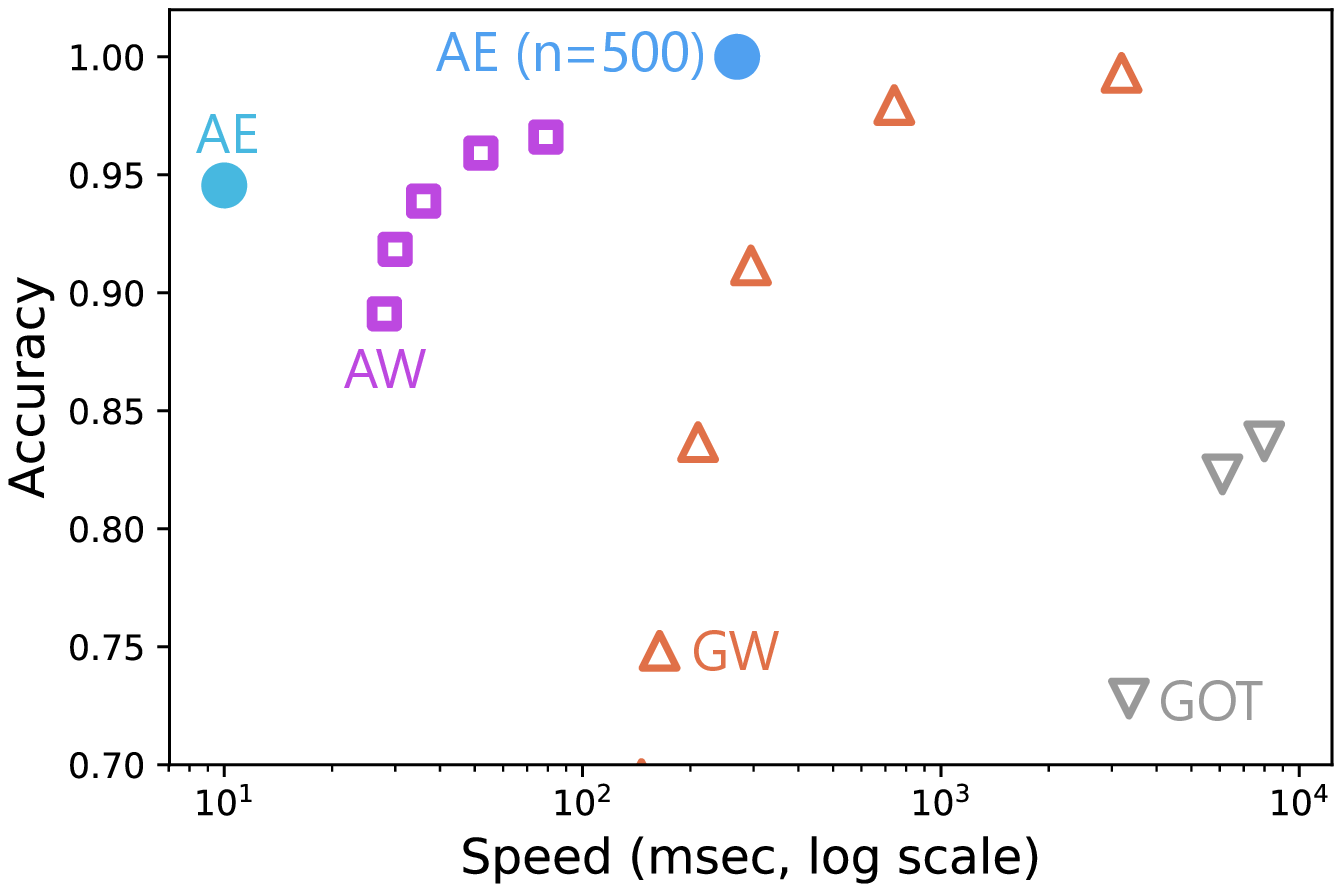}
  \end{center}
\caption{Top-$5$ accuracy of 1-NN.}
\label{fig: top5}
 \end{minipage}
\end{figure}

\subsection{Trade-off (Q2)}

In this experiment, we exclude the ``shark'' class from the dataset because this class contains only one shape. We first compute the geodesic distance matrix for each shape. For each shape, we then sample $n = 100$ points uniformly and randomly and sample the corresponding rows and columns of the distance matrix to make $\boldC \in \mathbb{R}_+^{n \times n}$. We compute the AE, AW, and GW distances of each pair of sampled distance matrices with uniform distribution. We compute distances using a large-scale computer cluster in parallel. The time consumption is the average time to compute the distance of a pair of shapes with a single core of Intel Xeon E7-4830 CPU for $100$ random pairs.

We use the official implementation of GOT \citep{maretic2019GOT} \url{https://github.com/Hermina/GOT} and vary the number of iterations to assess the trade-off in this experiment. We use negative exponential transformation $\boldW_{ij} = \exp(-\boldC_{ij}/\sigma)$ to construct the weight matrix $\boldW$ from the cost matrix $\boldC$. We use $\sigma = 10$ chosen from $\sigma = 5, 10, 20$ in validation. We also used the Gaussian kernel $\boldW_{ij} = \exp(-\boldC_{ij}^2/\sigma^2)$ but did not observe any improvement in early experiments. 

We also measure top-$3$ and top-$5$ accuracy of the $1$-NN classification. Figures \ref{fig: top3} and \ref{fig: top5} plot the Pareto optimal points of the accuracy and speed and show similar tendencies to Figure \ref{fig: shapes}. 

\subsection{Robustness (Q3)}

We set $\varepsilon = 10^{-6}$ for the AW distance, $\varepsilon = 10^{-8}$ for the robust AW distance, and $\varepsilon = 100$ for the GW distance. The regularization constants are different because they need to adjust to the scale of the ground costs used within AW, robust AW, and GW, which are themselves different. We found these hyperparameters did not affect the performance much in the early experiments.

\subsection{Assignment (Q4)}

We use the Barabasi Albert model with $n = 200$ nodes and $m = 2$ attachments in this experiment. The centrality-based baseline methods in Figure \ref{fig: matching} are as follows. Closeness: The closeness centrality \cite[\S 7.1.6]{newman2018networks}. Between: betweenness centrality \cite[\S 7.1.7]{newman2018networks}. Degree: degree centrality \cite[\S 7.1.1]{newman2018networks}. Eigenvector: eigenvector centrality \cite[\S 7.1.2]{newman2018networks}. Katz: Katz centrality \cite[\S 7.1.3]{newman2018networks}. PageRank: PageRank score \citep{PageRank}. HITS(Hub): hub score of the HITS algorithm \citep{HITS}. HITS(Auth): authority score of the HITS algorithm \citep{HITS}. We use the official implementation of GOT \citep{maretic2019GOT} \url{https://github.com/Hermina/GOT} and the official implementation of GWL \citep{graph-matching} \url{https://github.com/HongtengXu/gwl}. It should be noted that this task is more difficult than the task in the GWL paper \citep{graph-matching} because we match \emph{different} graphs (generated by the same model), whereas \cite{graph-matching} matched a graph with the \emph{same} graph with noise.

\subsection{Graph Model Comparison (Q5)}

We use the official implementation of GraphRNN \url{https://github.com/JiaxuanYou/graph-generation} for the GraphRNN, ER model, and BA model. The parameters of the ER model and BA model are estimated by the rule-based method (i.e., the ``general'' option in the implementation). Note that the authors of the original GraphRNN paper \citep{you2018graphrnn} used the exponential negative Wasserstein distance as the kernel of MMD, while we use the raw Wasserstein distance as the ground distance of the energy distance. The theory of the energy distance \citep{sejdinovic2013equivalence} shows that our energy distance can be interpreted as MMD and used for statistical hypothesis testing because the $1$-dimensional Wasserstein distance is conditionally negative.

\begin{figure}
    \centering
    \includegraphics[width=0.9\hsize]{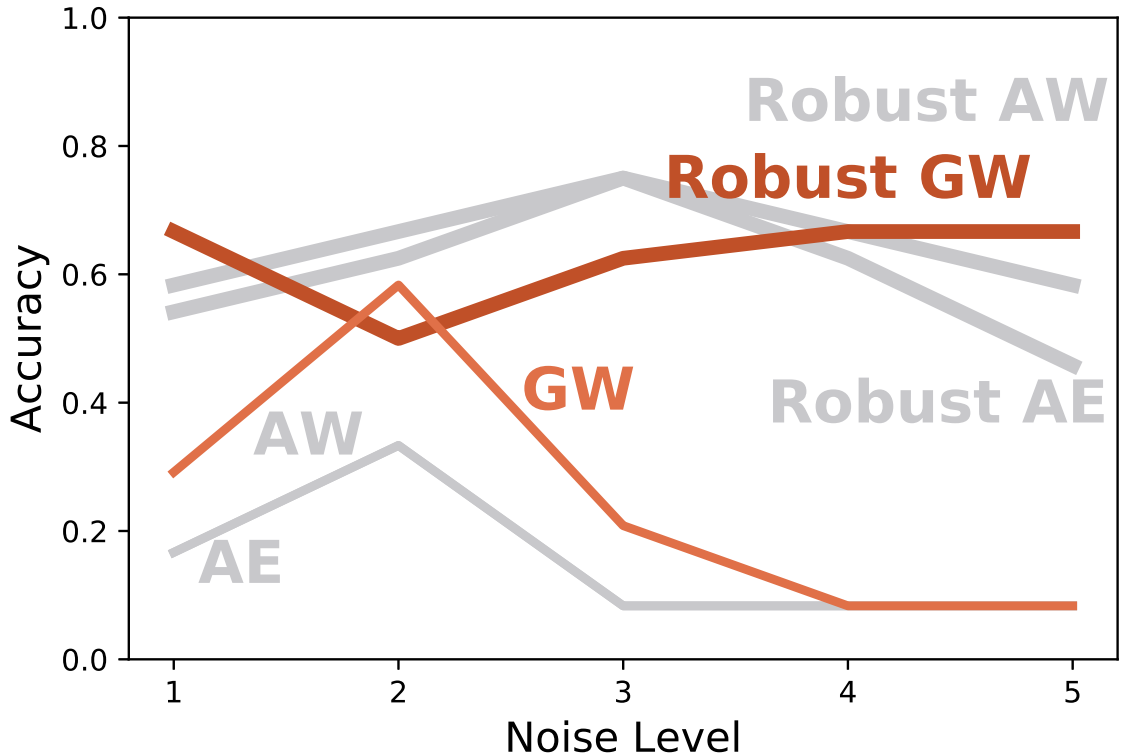}
    \caption{\textbf{Robust GW.} }
    \label{fig: robustGW}
\end{figure}

\begin{figure}
    \centering
    \includegraphics[width=0.9\hsize]{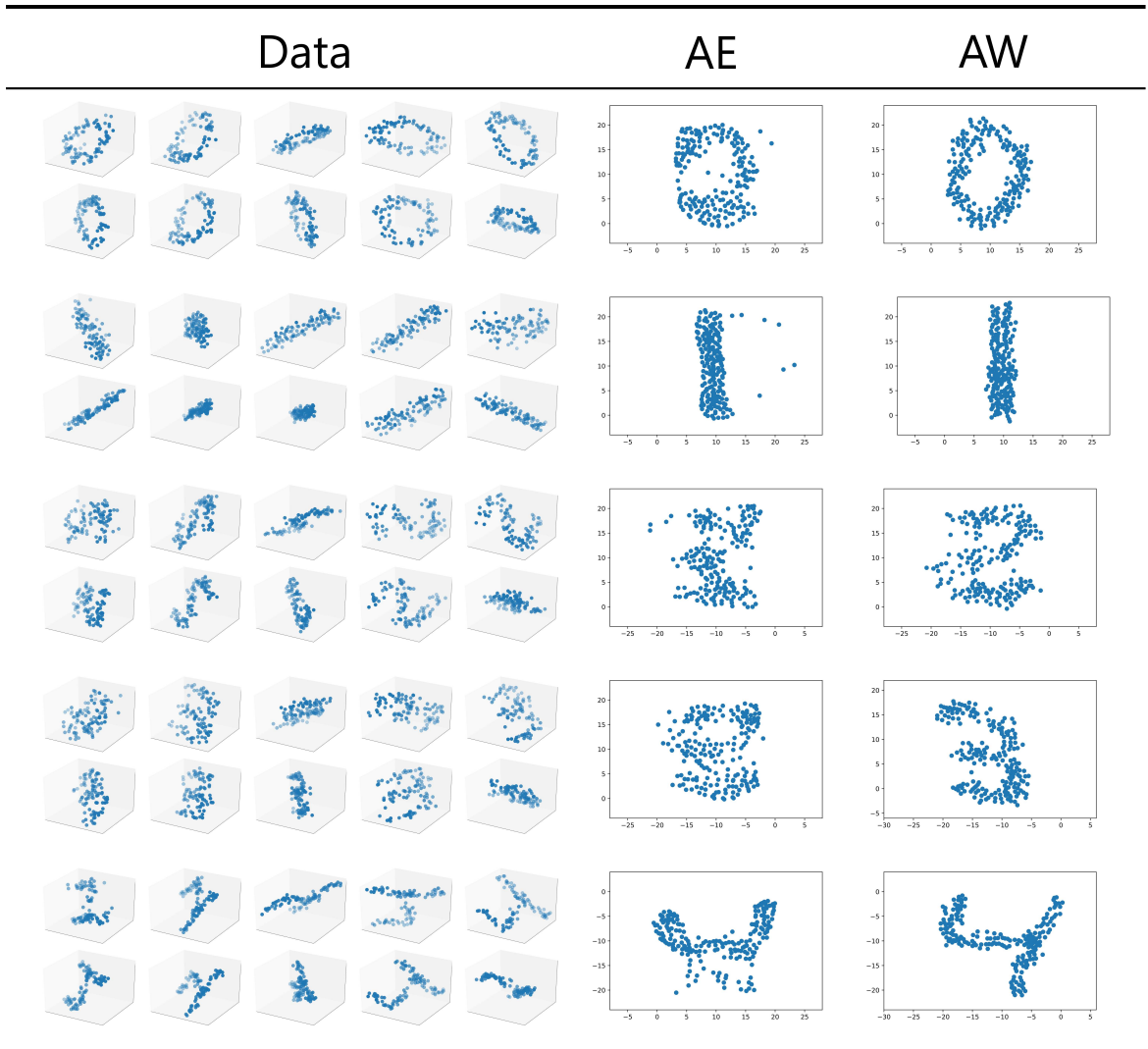}
    \caption{\textbf{Barycenters of 3D digits:} Barycenters of 3D digits in $2$-dimensional Euclidean space w.r.t. the AE and AW geometries. }
    \label{fig: barycenter}
\end{figure}

\section{Additional Experiments} \label{sec: barycenter}
\textbf{Robust GW: Can we apply our robustification technique to GW?} We apply our proposed robust technique (i.e., rank-based statistics) to the GW distance and conduct an experiments with the same setting as Q3. We would like to emphasize that applying the rank-based statistics to GW is also a part of our contribution. Figure \ref{fig: robustGW} shows that our technique makes the GW distance robust as well. Since the GW distance is used in many existing works, our proposed method is shown to have all the more significance in this field.

\textbf{Barycenter: Do AE and AW barycenters provide good summaries of datasets from different domains?} We compute barycenters of 3D shapes generated from the MNIST handwritten digits dataset \citep{MNIST} into $2$-dimensional Euclidean space with respect to the AE and AW geometries. It is difficult to summarize them because the data are lying in a different space (i.e., $3$-dimensional Euclidean space) from the barycenter ($2$-dimensional Euclidean space). The AE and AW distances can be applied to these settings since they can compare measures lying in different spaces. The shapes are rotated in $3$-dimensional spaces. Specifically, we generate the dataset by the following process: (1) we sample $100$ points $\{\boldp_i\}_{i = 1, \dots, 100}$ in the $2$-dimensional lattice $\{1, 2, \dots, 28\} \times \{1, 2, \dots, 28\}$ with a probability proportional to the brightness of a pixel, (2) we sample a random $3$-dimensional rotation matrix $\boldM$ uniformly, (3) we rotate points $\{\boldp_i\}$ by $\boldM$, and then (4) we add i.i.d. Gaussian noise $\mathcal{N}(\mathbf{0}, 0.25 \boldI)$ to each point. We optimize the sum of AE and AW distances from the data shapes by gradient descent. The gradients are computed by the auto-gradient package PyTorch \citep{PyTorch}. Figure \ref{fig: barycenter} shows the data and barycenters. The AE barycenters fails to provide good summaries of digits $2$ and $3$, whereas the AW barycenters preserve the characteristics of all digits thanks to more precise global matching.

\section{Proofs}

\begin{proofpn}{\ref{thm: variation}}
Since $H^{k}_i(x)$ is a piecewise-constant function,
\begin{align*}
\mathbb{E}_{h^1 \sim \mathcal{A}(S^1), h^2 \sim \mathcal{A}(S^2)}[\text{OT}_p^p(h^1, h^2)]
&= \sum_{i = 1}^{n} \sum_{j = 1}^{m} \bolda^1_i \bolda^2_j ~\text{OT}_p^p(\measure(i, \bolda^1, \boldC^1), \measure(j, \bolda^2, \boldC^2)) \\
&= \sum_{i = 1}^{n} \sum_{j = 1}^{m} \bolda^{1}_i \bolda^{2}_j \int_0^{\infty} |H^{1}_i(x) - H^{2}_j(x)| dx \\
&= \sum_{i = 1}^{n} \sum_{j = 1}^{m} \bolda^{1}_i \bolda^{2}_j \sum_{l = 1}^{K-1} (s_{l+1} - s_l) |H^{1}_i(s_l) - H^{2}_j(s_l)| \\
&= \sum_{l = 1}^{K-1} (s_{l+1} - s_l) \sum_{i = 1}^{n} \sum_{j = 1}^{m} \bolda^{1}_i \bolda^{2}_j |H^{1}_i(s_l) - H^{2}_j(s_l)| \\
&= \sum_{l = 1}^{K-1} (s_{l+1} - s_l) f(l).
\end{align*}
\end{proofpn}

\begin{proofpn}{\ref{thm: update}}
We assume $n_1 = n_2 = n$ without loss of generality by appropriately zero-padding $\bolda^1$ and $\bolda^2$.
\begin{align*}
f(l+1) - f(l)
&= -\left(\bolda^{k_l}_{i_l} \sum_{x = 1}^{n} \bolda^{k_l'}_{x}
|H^{k_l}_{i_l}(s_l) - H^{k_l'}_x(s_l)| \right) \\
& \quad +\left(\bolda^{k_l}_{i_l} \sum_{x = 1}^{n} \bolda^{k_l'}_{x}
|H^{k_l}_{i_l}(s_{l+1}) - H^{k_l'}_x(s_{l+1})| \right) \\
&= -\left(\bolda^{k_l}_{i_l} \sum_{x = 1}^{n} \bolda^{k_l'}_{x}
|c - H^{k_l'}_x(s_l)| \right) \\
& \quad +\left(\bolda^{k_l}_{i_l} \sum_{x = 1}^{n} \bolda^{k_l'}_{x}
|c' - H^{k_l'}_x(s_{l+1})| \right) \\
&= -\left(\bolda^{k_l}_{i_l} \sum_{x \colon H^{k'_l}_x(s_l) < c} \bolda^{k_l}_{x} c - \bolda^{k_l}_{x} H^{k'_l}_x(s_l) \right) \\
& \quad -\left(\bolda^{k_l}_{i_l} \sum_{x \colon c < H^{k'_l}_x(s_l)} \bolda^{k_l}_{x} H^{k'_l}_x(s_l) - \bolda^{k_l}_{x} c \right)\\
& \quad +\left(\bolda^{k_l}_{i_l} \sum_{x \colon H^{k'_l}_x(s_{l+1}) < c'} \bolda^{k_l}_{x} c - \bolda^{k_l}_{x} H^{k'_l}_x(s_{l+1}) \right) \\
& \quad +\left(\bolda^{k_l}_{i_l} \sum_{x \colon c' < H^{k'_l}_x(s_{l+1})} \bolda^{k_l}_{x} H^{k'_l}_x(s_{l+1}) - \bolda^{k_l}_{x} c\right) \\
&= -\bolda^{k_l}_{i_l} (\mathcal{S}_l^{k'}(-\infty, c) c - \mathcal{T}_l^{k'}(-\infty, c)) \\
& \quad -\bolda^{k_l}_{i_l} (\mathcal{T}_l^{k'}(c, \infty) - \mathcal{S}_l^{k'}(c, \infty) c) \\
& \quad +\bolda^{k_l}_{i_l} (\mathcal{S}_l^{k'}(-\infty, c') c' - \mathcal{T}_l^{k'}(-\infty, c')) \\
& \quad +\bolda^{k_l}_{i_l} (\mathcal{T}_l^{k'}(c', \infty) - \mathcal{S}_l^{k'}(c', \infty) c')
\end{align*}
\end{proofpn}

\end{document}